%% file: main.tex
\documentclass{article} %
\usepackage{iclr2021_conference,times}

\input{math_commands.tex}

\usepackage{hyperref}
\usepackage{url}

\title{Kanerva++: Extending the Kanerva Machine With
Differentiable, Locally Block Allocated Latent Memory}

\author{Jason Ramapuram$^{1,2}$, Yan Wu$^{3}$, Alexandros Kalousis$^{1}$ \\
  $^1$ University of Geneva \& Geneva School of Business Administration, HES-SO \\
  $^2$ Currently at Apple \\
  $^3$ Deepmind \\
  \texttt{jramapuram@apple.com, yanwu@google.com, alexandros.kalousis@hesge.ch}
}

\usepackage[none]{hyphenat} %
\usepackage{amssymb}
\usepackage{amsmath}
\usepackage{enumitem}
\usepackage{subcaption}
\usepackage{gensymb}
\usepackage{stmaryrd}
\usepackage{fancyhdr}
\usepackage{makecell}
\usepackage{float}
\usepackage{bm}
\usepackage{bbm} %
\usepackage{relsize}
\usepackage{graphicx} %
\usepackage{adjustbox}
\usepackage{algorithm2e}

\iclrfinalcopy %
\begin{document}

\maketitle

\vskip -0.2in
\begin{abstract}
\vskip -0.15in
Episodic and semantic memory are critical components of the human memory model.
The theory of complementary learning systems \citep{mcclelland1995there} suggests
that the compressed representation produced by a serial event (episodic memory)
is later restructured to build a more generalized form of reusable knowledge
(semantic memory).
In this work we develop a new principled Bayesian memory allocation scheme that %
bridges the gap between episodic and semantic memory via a hierarchical latent variable model.  We take
inspiration from traditional heap allocation and extend the idea of
locally contiguous memory to the Kanerva Machine, enabling a novel
differentiable block allocated latent memory. In contrast to the Kanerva Machine, we simplify the process of memory writing by
treating it as a fully feed forward deterministic process, relying on the
stochasticity of the read key distribution to disperse information
within the memory. We demonstrate that this allocation scheme improves
performance in \emph{memory conditional} image generation, resulting
in new state-of-the-art
conditional likelihood values on binarized MNIST
(\textbf{$\leq$41.58 nats/image})
, binarized Omniglot
(\textbf{$\leq$66.24 nats/image}), as well
as presenting competitive performance on CIFAR10, DMLab Mazes, Celeb-A and ImageNet32$\times$32.
\end{abstract}
\vspace{-0.1in}

\input{introduction}

\input{related}

\input{model}

\input{experiments}

\vspace{-0.1in}
\section{Conclusion}
\vspace{-0.1in}
In this work, we propose a novel block allocated memory in a generative
model framework and demonstrate its state-of-the-art performance
on several memory conditional image generation
tasks. We also show that stochasticity in low-dimensional spaces
produces higher quality samples in comparison to high-dimensional
latents typically used in VAEs.
Furthermore, perturbations to the low-dimensional
key generate samples with high variations. Nonetheless, there are still many unanswered questions: would a hard
attention based solution to differentiable indexing prove to be better
than a spatial transformer? What is the optimal upper bound of window
read regions based on the input distribution? Future work will
hopefully be able to address these lingering issues and %
further improve generative memory models.

\bibliography{bibliography}
\bibliographystyle{iclr2021_conference}

\newpage
\input{appendix}

\end{document}

%% file: math_commands.tex
\usepackage{amsmath,amsfonts,bm}

\def\eqref#1{equation~\ref{#1}}

\def\1{\bm{1}}

\DeclareMathAlphabet{\mathsfit}{\encodingdefault}{\sfdefault}{m}{sl}
\SetMathAlphabet{\mathsfit}{bold}{\encodingdefault}{\sfdefault}{bx}{n}

%% file: introduction.tex
\vspace{-0.1in}
\section{Introduction}
\vspace{-0.1in}

Memory is a central tenet in the model of human intelligence and is crucial to
long-term reasoning and planning. %
Of particular interest is the theory of complementary learning systems \cite{mcclelland1995there} which proposes that the brain employs two complementary systems to support the
acquisition of complex behaviours: a hippocampal fast-learning system that
records events as episodic memory, and a neocortical slow learning system that
learns statistics across events as semantic memory. While the functional
dichotomy of the complementary systems are well-established
\cite{mcclelland1995there,kumaran2016learning}, it remains unclear whether they
are bounded by different computational principles. In this work we introduce a
model that bridges this gap by showing that the same statistical learning
principle can be applied to the fast learning system through the construction of
a hierarchical Bayesian memory.

\vskip -0.1in
\begin{figure}[H]
  \begin{center}
\begin{minipage}{0.5\textwidth}
  \begin{center}
    \centerline{\includegraphics[width=\linewidth]{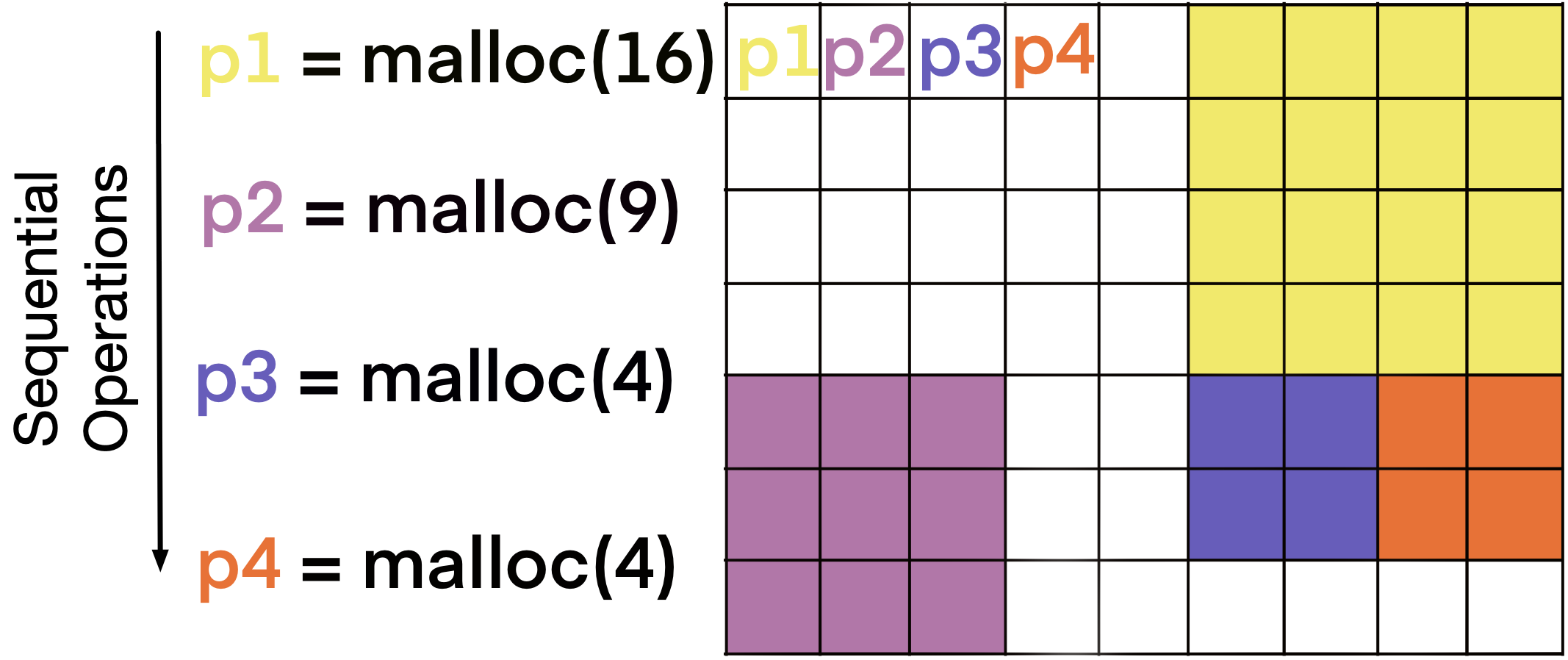}}
  \end{center}
\end{minipage}%
\begin{minipage}{0.5\textwidth}
  \begin{center}
    \centerline{\includegraphics[width=0.65\linewidth]{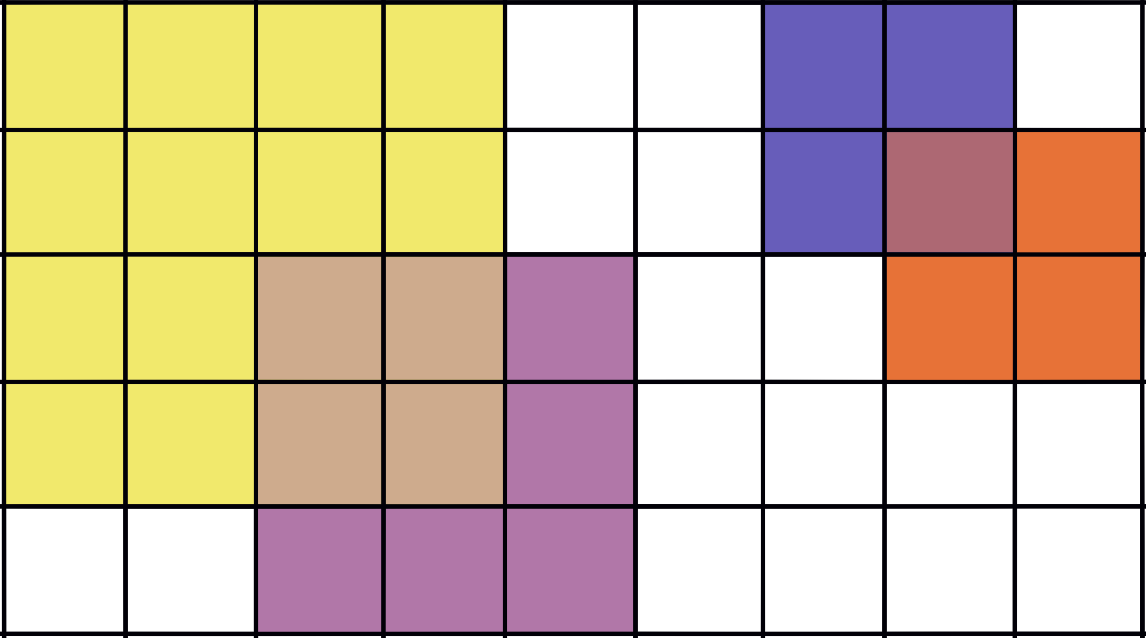}}
    \end{center}
  \end{minipage}%
\end{center}
\vskip -0.25in
\caption{Example final state of a traditional heap allocator
\citep{marlow2008parallel} (\emph{Left}) vs. K++ (\emph{Right}); final state
created by sequential operations listed on the left. K++ uses a key distribution
to stochastically point to a memory sub-region while \citet{marlow2008parallel}
uses a direct pointer. Traditional heap allocated memory affords $O(1)$ free /
malloc computational complexity and serves as inspiration for K++ which uses differentiable neural proxies.
}\label{allocator}
\end{figure}
\vskip -0.2in

While recent work has shown that using memory augmented neural networks can
drastically improve the performance of generative models
\citep{wu2018kanerva,wu2018learning}, language models
\citep{DBLP:journals/corr/WestonCB14}, meta-learning
\citep{DBLP:conf/icml/SantoroBBWL16}, long-term planning
\citep{graves2014neural,graves2016hybrid} and sample efficiency in reinforcement
learning \citep{zhu2019episodic}, no model has been proposed to exploit the
inherent multi-dimensionality of biological memory \cite{10.3389/fncom.2017.00048}. Inspired by the traditional
(computer-science) memory model of heap allocation (Figure
\ref{allocator}-\emph{Left}), we propose a novel differentiable memory
allocation scheme called Kanerva ++ (K++), that learns to compress an episode of
samples, referred to by the set of pointers $\{p1, ..., p4\}$ in Figure
\ref{allocator}, into a latent multi-dimensional memory (Figure
\ref{allocator}-\emph{Right}). The K++ model infers a key distribution as a
proxy to the pointers \citep{marlow2008parallel} and
is able to embed similar samples to an overlapping latent representation space,
thus enabling it to be more efficient on compressing input distributions.
In this work, we focus on applying this novel memory allocation scheme to
latent variable generative models, where we improve the memory model in the Kanerva
Machine \citep{wu2018kanerva,wu2018learning}.

%% file: related.tex
\vspace{-0.2in}
\section{Related Work}
\vspace{-0.1in}

\textbf{Variational Autoencoders}:\  Variational autoencoders (VAEs) \citep{kingma2014} are a fundamental part of
the modern machine learning toolbox and have wide ranging applications
from generative modeling
\citep{kingma2014,kingma2016improved,DBLP:journals/corr/BurdaGS15},
learning graphs \citep{DBLP:journals/corr/KipfW16a}, medical applications \citep{sedai2017semi,zhao2019variational} and video
analysis \citep{fan2020video}. %
As a latent variable model, VAEs infer an
approximate posterior over a latent representation $Z$
and can be used in downstream tasks such as control in reinforcement learning
\citep{DBLP:conf/nips/NairPDBLL18,DBLP:conf/icml/PritzelUSBVHWB17}. VAEs
maximize an evidence lower bound (ELBO), $\mathcal{L}(X, Z)$, of the
log-marginal likelihood, $\ln p(X) > \mathcal{L}(X, Z) = \ln p_{\theta}(X|Z) -
  \mathcal{D}_{KL}(q_{\phi}(Z|X) || p_{\theta}(Z))$.
The produced variational approximation, $q_{\phi}(Z|X)$, is typically called the encoder, while $p_{\theta}(X|Z)$ %
comes from the decoder.
Methods that aim to improve these latent variable generative models typically fall
into two different paradigms: learning more informative priors or
leveraging novel decoders. While improved decoder models such as PixelVAE
\citep{DBLP:conf/iclr/GulrajaniKATVVC17} and PixelVAE++ \citep{DBLP:journals/corr/abs-1908-09948}
drastically improve the performance of $p_{\theta}(X|Z)$, they suffer
from a phenomenon called posterior collapse
\citep{DBLP:conf/iclr/LucasTGN19}, where the decoder can become almost
independent of the posterior sample, but still retains
the ability to reconstruct the original sample by relying on its auto-regressive property
\citep{goyal2017z}.

In contrast, VampPrior \citep{tomczak2018vae}, Associative Compression Networks (ACN)
\citep{graves2018associative}, VAE-nCRP \citep{goyal2017nonparametric}
and VLAE \citep{chen2017variational} tighten the variational bound by learning more
informed priors. VLAE for example, uses a powerful auto-regressive prior; VAE-nCRP learns
a non-parametric Chinese restaurant process prior and VampPrior learns
a Gaussian mixture prior representing prototypical virtual
samples. %
On the other hand, ACN takes a two-stage approach: by clustering real samples in
the space of the posterior; and by using these related samples as inputs
to a learned prior, ACN provides an information theoretic alternative
to improved code transmission. Our work falls into this latter
paradigm: we parameterize a learned prior by reading from a common
memory, built through a transformation of an episode of input samples.

\textbf{Memory Models}:\ Inspired by the associative nature of biological memory,
the Hopfield network \citep{hopfield1982neural} introduced the notion of
content-addressable memory, defined by a set of binary neurons coupled with a
Hamiltonian and a dynamical update rule. Iterating the update rule minimizes the
Hamiltonian, resulting in patterns being stored at different configurations
\citep{hopfield1982neural,krotov2016dense}. Writing in a Hopfield network, thus
corresponds to finding weight configurations such that stored patterns become
attractors via Hebbian rules \citep{hebb1949organization}. This concept of
memory was extended to a distributed, continuous setting in
\citet{kanerva1988sparse} and to a complex valued, holographic convolutional
binding mechanism by \citet{plate1995holographic}. The central difference
between associative memory models \cite{hopfield1982neural,kanerva1988sparse}
and holographic memory \cite{plate1995holographic} is that the latter decouples
the size of the memory from the input word size.

Most recent work with memory augmented neural networks treat memory in a
slot-based manner (closer to the associative memory paradigm), where each column
of a memory matrix, $M$, represents a single slot. Reading memory traces, $z$,
entails using a vector of addressing weights, $r$, to attend to the appropriate
column of $M$, $z = r^TM$. This paradigm of memory includes models such as the
Neural Turing Machine (NTM) \citep{graves2014neural}, Differentiable Neural
Computer (DNC) \citep{graves2016hybrid} \footnote{While DNC is slot based, it
should be noted that DNC reads rows rather than columns.}, Memory Networks
\citep{DBLP:journals/corr/WestonCB14}, Generative Temporal Models with Memory
(GTMM) \cite{DBLP:conf/icml/FraccaroRZPEV18}, Variational Memory Encoder-Decoder
(VMED) \cite{le2018variational}, and Variational Memory Addressing (VMA)
\citep{bornschein2017variational}. VMA differs from GTMM, VMED, DNC, NTM and
Memory Networks by taking a stochastic approach to discrete key-addressing,
instead of the deterministic approach of the latter models.

Recently, the Kanerva Machine (KM) \citep{wu2018kanerva} and its extension,
the Dynamic Kanerva Machine (DKM) \citep{wu2018learning}, interpreted
memory writes and reads as inference in a generative model, wherein
memory is now treated as a distribution, $p(M)$.
Under this framework, memory reads and writes are recast as sampling
or updating the memory posterior. The DKM model differs from the KM
model by introducing a dynamical addressing rule that could be used
throughout training.
While providing an intuitive and theoretically sound bound on the data
likelihood, the DKM model requires an inner optimization loop which entails solving an ordinary
least squares (OLS) problem. Typical OLS solutions require a matrix
inversion ($O(n^3)$), preventing the model from scaling to large
memory sizes. More recent work has focused on employing a product of smaller
Kanerva memories \citep{marblestone2020product} in an effort to
minimize the computational cost of the matrix inversion.
In contrast, we propose a simplified view of memory creation by
treating memory writes as a deterministic process in a fully feed-forward
setting. Crucially, we also modify the read operand such that it uses localized
sub-regions of the memory, providing an extra dimension of operation in
comparison with the KM and DKM models. While the removal of memory stochasticity
might be interpreted as reducing the representation power of the model, we
empirically demonstrate through our experiments that our model performs better, trains
quicker and is simpler to optimize. The choice of a deterministic memory is
further reinforced by research in psychology, where human visual memory has been
shown to change deterministically \cite{gold2005visual,spencer2002prototypes, hollingworth2013visual}.

%% file: model.tex
\section{Model}

To better understand the K++ model, we examine each of the individual
components to understand their role within the complete
generative model. We begin by first deriving a conditional variational lower
bound (Section \ref{vlb}), describing the optimization objective and
probabilistic assumptions. We then describe the write operand (Section
\ref{wm}), the generative process (Section \ref{sgm}) and finally the read and iterative read operands (Section \ref{rm}).

\subsection{Preliminaries} \label{vlb}

K++ operates over an exchangeable episode
\citep{aldous1985exchangeability} of samples, $X = \{x_t\}_{t=1}^T \in \mathcal{D}$, drawn from a dataset
$\mathcal{D}$, as in the Kanerva Machine. Therefore, the ordering of the samples
within the episode does not matter. This enables factoring the
conditional, $p(X|M)$, over each of the individual samples:
$\prod_{t=1}^T p(x_t|M)$, given the memory, $M \sim p(M), M \in
\mathbb{R}^{\hat{C} \times \hat{W} \times \hat{H}}$. Our objective in this work %
is to maximize the expected conditional log-likelihood as in
\citep{bornschein2017variational,wu2018kanerva}:

\vskip -0.3in
\begin{align}
  \mathcal{J} = \mathbb{E}_{p(X),\ p(M|X)} \ln p_{\theta} (X|M) =
  \int \int p(X) p(M|X) \sum_{t=1}^T \ln p_{\theta}(x_t|M) dX
  dM \label{cond_indep}
\end{align}
As alluded to in \citet{barber2004information} and
\citet{wu2018kanerva}, this objective can be interpreted as maximizing
the mutual information, $I(X; M)$, between the memory, $M$, and the episode, $X$, since $I(X; M) = H(X) +
\mathcal{J} = H(X) - H(X|M)$ and given that the entropy of the data, $H(X)$, is
constant. In order to actualize Equation \ref{cond_indep} we rely on a variational bound which we
derive in the following section.

\subsection{Variational Lower Bound}
\input{vlb_with_z_v2}

\vspace{-0.1in}
\subsection{Write Model} \label{wm}
\begin{figure}[H]
\begin{minipage}{0.69\textwidth}
  \includegraphics[width=\linewidth]{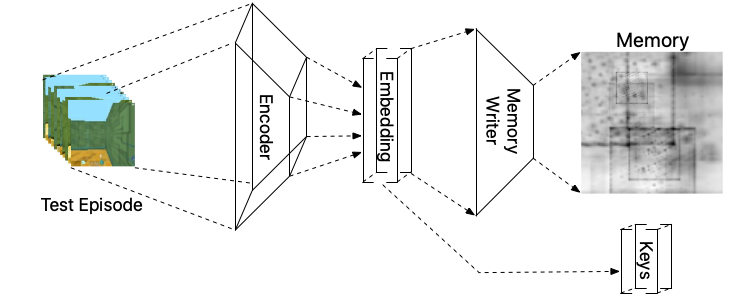}
\end{minipage}%
\hfill\vline\hfill
  \begin{minipage}{0.30\textwidth}
    \begin{algorithm}[H]
      \small
sample episode: $X = \{x_1, ... , x_T\} \sim \mathcal{D}$\;
compute embedding: $E = f_{\theta_{enc}}(X)$ \;
infer keys: $Y \sim \mathcal{N}(\mu_{\theta_{key}}(E), \sigma^2_{\theta_{key}}(E)$) \;
write memory: $M \sim \delta(f_{\theta_{mem}}(E))$
\vskip 0.2in
\end{algorithm}
\end{minipage}%
\caption{\emph{Left}: Write model. \emph{Right}: Write operation.} \label{write_model}
\end{figure}

Writing to memory in the K++ model (Figure \ref{write_model}) entails encoding
the input episode, $X = \{x_t\}_{t=1}^T$, through the encoder,
$E = f_{\theta_{enc}}(X)$, pooling the representation over the episode and
encoding the pooled representation with the memory writer,
$M = f_{\theta_{mem}}(E)$. In this work, we employ a Temporal Shift Module (TSM)
\citep{lin2019tsm}, applied on a ResNet18 \citep{he2016deep}. TSM works by
shifting feature maps of a two-dimensional vision model in the temporal dimension in order to
build richer representations of contextual features. In the case of K++, this
allows the encoder to build a better representation of the memory by leveraging
intermediary episode specific features. Using a TSM encoder over a standard
convolutional stack improves the performance of both K++ and DKM, where the
latter observes an improvement of 6.32 nats/image over the reported test conditional
variational lower bound of 77.2 nats/image \citep{wu2018learning} for the
binarized Omniglot dataset. As far as we are aware, the application of a TSM encoder to
memory models has not been explored and is a contribution of this work.

The memory writer model, $f_{\theta_{mem}}$, in Figure
\ref{write_model}, allows K++ to non-linearly transform the pooled
embedding to better summarize the episode. In addition to inferring
the deterministic memory, $M$, we also project the non-pooled
embedding, $E$, through a key model, $f_{\theta_{key}}$:

\vskip -0.2in
\begin{align}
  Y = \mu_{\theta_{key}}(E) + \sigma^2_{\theta_{key}}(E) \odot \epsilon,\
  \ \ y_{tk} \in \mathbb{R}^{3},\ \ \ \epsilon \sim \mathcal{N}(0, 1).
\end{align}

The reparameterized keys will be used to read sample specific memory traces, $\hat{M}$, from the full memory, $M$. The memory traces, $\hat{M}$, are used in training through the
learned prior, $p_{\theta}(Z|\hat{M}, Y) = \mathcal{N}(\mu_{z}, \sigma^2_z)$,
from Equation \ref{actual_elbo} via the KL divergence,
$\mathbb{E}_{q_{\phi}(Y|X)}\mathcal{D}_{KL}[q_{\phi}(Z|X) || P_{\theta}(Z|\hat{M}, Y)]$. This KL divergence constrains the optimization objective to keep the representation of the
amortized approximate posterior, $q_{\phi}(Z|X)$, (probabilistically) close to
the memory readout representation of the learned prior,
$p_{\theta}(Z|\hat{M}, Y)$. In the generative setting, this constraint enables memory traces to be
routed from the learned prior, $p_{\theta}(Z|\hat{M}, Y)$, to the decoder, $p_{\theta}(X|\cdot)$, in a similar manner to standard VAEs. We detail this process in the following section.

\vspace{-0.1in}
\subsection{Sample Generation} \label{sgm}

\begin{figure}[H]
\begin{minipage}{0.71\textwidth}
  \includegraphics[width=\linewidth]{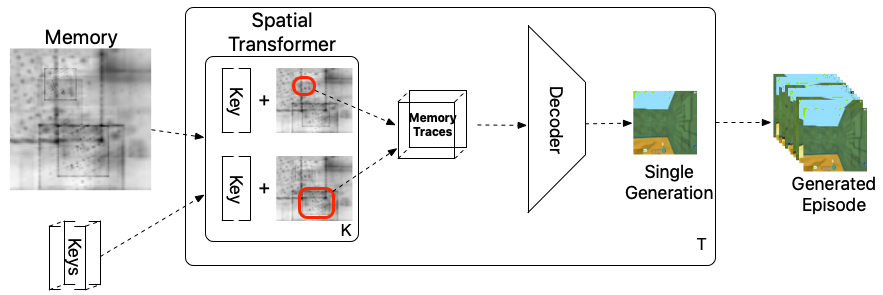}
\end{minipage}%
\hfill\vline\hfill
  \begin{minipage}{0.28\textwidth}
    \begin{algorithm}[H]
      \small
      given memory: $M$ \;
      sample keys: $Y \sim P(Y)$\;
      extract regions: $\hat{M} = \{f_{\theta_{ST}}(M,
      y_{tk})\}_{k=1}^K$\;
      infer latent: $Z \sim \mathcal{N}(\mu_{\theta_{Z}}(\hat{M}),
      \sigma^2_{\theta_{Z}}(\hat{M}))$ \;
      decode: $\hat{X}_t \sim
      \mathcal{N}(\mu_{\theta_{dec}}(\mu_{Z}), \sigma^2)$
\vskip 0.2in
\end{algorithm}
\end{minipage}%
\caption{\emph{Left}: Generative model. \emph{Right}: Generative operation.}
\end{figure}

The Kanerva++ model, like the original KM and DKM models, enables sample
generation given an existing memory or set of memories. $K$ samples from the
prior key distribution,
$\{y_k\}_{k=1}^K \sim p(Y) = \mathcal{N}(0, 1), y_k \in \mathrm{R}^{3}$, are
used to parameterize the spatial transformer, $f_{ST}$, which indexes the
deterministic memory, $M$. The result of this differentiable indexing is a set
of memory sub-regions, $\hat{M}$, which are used in the decoder,
$p_{\theta}(X|\cdot)$, to generate synthetic samples. Reading samples in this
manner forces the encoder to utilize memory sub-regions that are useful for
reconstruction, as non-read memory regions receive zero gradients during
backpropagation. This insight allows us to use the simple feed-foward write process
described in Section \ref{wm}, while still retaining the ability to produce locally
contiguous block allocated memory.

\subsection{Read / Iterative Read Model} \label{rm}
\vspace{-0.1in}

\begin{figure}[H]
  \begin{center}
\scalebox{1.0}{\parbox{1.0\linewidth}{
\begin{minipage}{0.75\textwidth}
  \includegraphics[width=\linewidth]{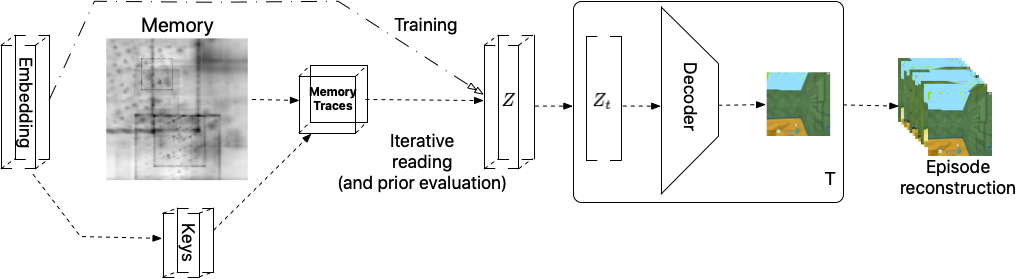}
\end{minipage}%
\hfill\vline\hfill
\begin{minipage}{0.24\textwidth}
    \begin{algorithm}[H]
      \small
      given embedding: $E$ \;
      \eIf{training}{
        infer latent: $Z \sim \mathcal{N}(\mu_{\theta_{z}}(E), \sigma^2_{\theta_{z}}(E))$\\
        infer prior: $Z \sim \mathcal{N}(\mu_{\theta_{z}}(\hat{M}), \sigma^2_{\theta_{z}}(\hat{M}))$
      }{
      infer latent: $Z \sim \mathcal{N}(\mu_{\theta_{z}}(\hat{M}), \sigma^2_{\theta_{z}}(\hat{M}))$
      }
      decode: $\hat{X} \sim \mathcal{N}(\mu_{\theta_{dec}}(\mu_{Z}), \sigma^2)$
    \end{algorithm}
  \end{minipage}%
}}
\end{center}
\caption{\emph{Left}: Read model: bottom branch from embedding used during iterative reading and prior evaluation. Stable top branch used to infer $q_{\phi}(Z|X)$ during training. \emph{Right}: Read operation.} \label{read_model}
\end{figure}

K++ involves two forms of reading (Figure \ref{read_model}): iterative reading
and a simpler and more stable read model used for training. During training we
actualize $q_{\phi}(Z|X)$ from Equation \ref{actual_elbo} using an amortized
isotropic-gaussian posterior that directly transforms the embedding of the
episode, $E$, using a learned neural network (Figure
\ref{plate}-\emph{b}). As mentioned in Section \ref{rm}, the readout,
$Z$, of the memory traces, $\hat{M}$, are encouraged to learn a meaningful
structured representation through the memory read-out KL divergence,
$\mathbb{E}_{q_{\phi}(Y|X)}\mathcal{D}_{KL}[q_{\phi}(Z|X) || P_{\theta}(Z|\hat{M}, Y)]$,
which attempts to minimize the (probabilistic) distance between $q_{\phi}(Z|X)$ and $P_{\theta}(Z|\hat{M}, Y)$.

Kanerva memory models also possess the ability to gradually improve a sample
through interative inference (Figure \ref{plate}-\emph{c}), whereby noisy samples can be improved by leveraging
contextual information stored in memory. This can be interpreted
as approximating the posterior, $q(Z |X, M)$, by marginalizing the approximate key distribution:
\begin{align}
  q(Z |X, \hat{M}) &= \int q_{\phi}(Y|X) p_{\theta}(Z | Y, \hat{M}) \delta Y \approx p_{\theta}(Z | Y=Y^*, \hat{M}), \label{full_posterior} %
\end{align}
where $Y^* \sim q_{\phi}(Y|X=\hat{X})$ in Equation \ref{full_posterior} uses a single sample Monte Carlo
estimate, evaluated by re-infering the previous reconstruction, $\hat{X} \sim p_{\theta}(X|\cdot)$, through
the approximate key posterior. Each subsequent memory readout, $Z$, improves upon its
previous representation by absorbing additional information from the memory.

%% file: vlb_with_z_v2.tex
To efficiently read from the memory, $M$, we introduce a set
of latent variables corresponding to the $K$ addressing read heads,
$Y = \{\{y_{tk}\}_{k=1}^K\}_{t=1}^T, y_{tk} \in \mathbb{R}^3$
, and a set of latent variables
corresponding to the readout from the memory, $Z =
\{z_{t}\}_{t=1}^T, z_t \in \mathbb{R}^L$.
Given these latent variables, we can decompose the conditional, $\ln p(X|M)$, using the
product rule
and introduce variational approximations $q_{\phi}(Z|X)$ \footnote{We use $q_{\phi}(Z|X)$ as our variational approximation instead of $q_{\phi}(Z|X, Y, M)$ in DKM as it presents a more stable objective. We discuss this in more detail in Section \ref{rm}.} and $q_{\phi}(Y|X)$
via a multipy-by-one trick:

\vspace{-0.2in}
\begin{align}
  \ln p(X|M) &= \ln \frac{p(X, Z, Y|M)}{p(Z, Y| X, M)} = \ln
             \frac{p(X|Z, M, Y)\ p(Z|M, Y)\ p(Y | M)}{p(Z|M, Y, X)\
               p(Y|M, X)} \nonumber \\
           &\gtrapprox \mathbb{E}_{q_{\phi}(Z | X), q_{\phi}(Y | X)}\bigg(\ln \frac{p(X|Z, M, Y)\ p(Z|M, Y)\ p(Y)}{p(Z|M,
             Y, X)\ p(Y|M, X)\ q_{\phi}(Z | X)\ q_{\phi}(Y|X)}\bigg) \label{eqn2} \\
           &= \mathcal{L}_T + \underbrace{\mathbb{E}_{q_{\phi}(Y|X)}
             \mathcal{D}_{KL}(q_{\phi}(Z|X) || p(Z|M, Y, X))}_{\geq
             0} \label{full_elbo}\\
           &\hspace{.37in}+\underbrace{\mathcal{D}_{KL}(q_{\phi}(Y|X)
             || p(Y|M, X))}_{\geq 0} \nonumber
\end{align}
\vskip -0.13in
Equation \ref{eqn2} assumes that $Y$ is independent from $M$: $p(Y|M) = p(Y)$.
This decomposition results in Equation \ref{full_elbo}, which
includes two KL-divergences
against true (unknown) posteriors, $p(Z| M, Y, X)$ and $p(Y|M, X)$.
We can then train the model by maximizing the evidence lower bound
(ELBO), $\mathcal{L}_T$, to the true conditional, $\ln p(X | M) > \mathcal{L}_T$:
\vskip -0.2in

\begin{align}
  \mathcal{L}_T &= \underbrace{\mathbb{E}_{q_{\phi}(Z|X), \ q_{\phi}(Y|X)}
                  \ln p_{\theta}(X|Z, M, Y)}_{\text{Decoder}}\nonumber \\
                &\hspace{0.1in} -\underbrace{\mathbb{E}_{q_{\phi}(Y|X)}\mathcal{D}_{KL}[q_{\phi}(Z|X) ||
                  p_{\theta}(Z|M, Y)]}_{\text{Amortized latent
                  variable posterior vs. memory readout
prior}} \label{actual_elbo}\\
                &\hspace{0.1in} -\underbrace{\mathcal{D}_{KL}[q_{\phi}(Y|X)
                  || p(Y)]}_{\text{Amortized key posterior vs. key prior}}\nonumber
  \end{align}
\vskip -0.1in
The bound in Equation \ref{actual_elbo} is tight %
if $q_{\phi}(Z|X) = p(Z|M, Y, X)$ and $q_{\phi}(Y|X) = p(Y | M, X)$, however, %
it involves inferring the
entire memory $M \sim q_{\phi}(M|X, Y)$. This prevents us from decoupling the size of the memory from
inference and scales the computation complexity based on the size of the
memory. To alleviate this constraint, we assume a purely deterministic memory,
$M \sim \delta[f_{mem}(f_{enc}(X))]$, built by transforming the input episode,
$X$, via a deterministic encoder and memory transformation model, $f_{mem} \circ f_{enc}$. We
also assume that the regions of memory which are useful in reconstructing a
sample, $x_t$, can be summarized by a set of $K$ localized contiguous memory
sub-blocks as described in Equation \ref{memapprox} below. The intuition here is
that similar samples, $x_{t} \approx x_{r}$, might occupy a disjoint part of the
representation space and the decoder, $p_{\theta}(X|\cdot)$, would need to read multiple regions to properly handle sample reconstruction. For example, the digit ``3'' might share part of the representation space with a ``2'' and another part with a ``5''.
\vskip -0.23in
\begin{align}
\hat{M} \sim \delta\bigg[\{m_k = f_{ST}(M=f_{mem}(f_{enc}(X)), y_{tk})\}_{k=1}^K\bigg] \approx q_{\phi}(M | X, Y) \label{memapprox}
\end{align}
\vskip -0.13in

$\hat{M}$ in equation \ref{memapprox} represents a set of $K$ dirac-delta memory
sub-regions, determined by the addressing key, $y_{tk}$, and a spatial
transformer ($ST$) network \cite{jaderberg2015spatial}, $f_{ST}$ \footnote{We
provide a brief review of spatial transformers in Appendix \ref{streview}}. Our
final optimization objective, $\mathcal{L}_T$, is attained by approximating
$M \sim q_{\phi}(M | X, Y)$ from Equation \ref{actual_elbo} with $\hat{M}$ (Equation \ref{memapprox}) and is
summarized by the graphical model in \ref{plate} below.

\vskip -0.13in
\begin{figure}[H]
  \centerline{\includegraphics[width=0.9\linewidth]{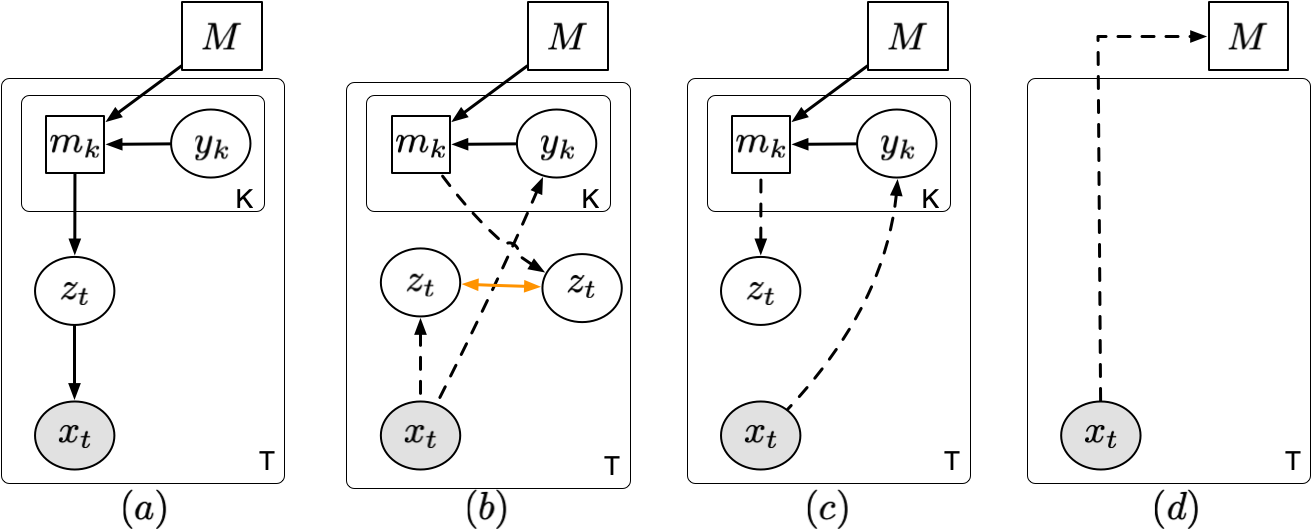}}
  \caption{\emph{(a)}: Generative model (\S \ref{sgm}). \emph{(b)}: Read
inference model (\S \ref{rm}). \emph{(c)}: Iterative read inference model (\S
\ref{rm}). \emph{(d)}: Write inference model (\S \ref{wm}). Dashed lines
represent approximate inference, while solid lines represent computing of a
conditional distribution. Double sided arrow in (\emph{b}) represents the KL
divergence between $q_{\phi}(Z|X)$ and $p_{\theta}(Z|M, Y)$ from Equation
\ref{actual_elbo}. Squares represent deterministic nodes. Standard plate
notation is used to depict a repetitive operation.}\label{plate}
\end{figure}

%% file: experiments.tex
\section{Experiments}

We contrast K++ against state-of-the-art memory conditional vision models %
and present empirical results in Table \ref{likelihood_table}.  Binarized
datasets assume Bernoulli output distributions, while continuous values are
modeled by a discretized mixture of logistics
\cite{DBLP:conf/iclr/SalimansK0K17}. As is standard in literature \cite{DBLP:journals/corr/BurdaGS15,DBLP:journals/corr/abs-1908-09948,ma2018mae,chen2017variational}, we provide
results for binarized MNIST and binarized Omniglot in nats/image and rescale the corresponding results to bits/dim for all other datasets.
We describe the model architecture, the optimization procedure and the memory
creation protocol in Appendix \ref{appendix_arch} and \ref{appendix_exp}.
Finally, extra Celeb-A generations and test image reconstructions for all
experiments are provided in Appendix \ref{celeba_gen_sec} and Appendix
\ref{reconstructions} respectively.

\begin{table}[H]
  \scalebox{0.9}{\parbox{1.0\linewidth}{%
\begin{tabular}{llllll}
\multicolumn{1}{c}{\textbf{Method}} & \multicolumn{1}{c}{\textbf{\begin{tabular}[c]{@{}c@{}}Binarized\\ MNIST\\ (nats / image)\end{tabular}}} & \multicolumn{1}{c}{\textbf{\begin{tabular}[c]{@{}c@{}}Binarized\\ Omniglot\\ (nats / image)\end{tabular}}} & \multicolumn{1}{c}{\textbf{\begin{tabular}[c]{@{}c@{}}Fashion\\ MNIST\\ (bits / dim)\end{tabular}}} & \textbf{\begin{tabular}[c]{@{}l@{}}CIFAR10\\ \\ (bits / dim)\end{tabular}} & \multicolumn{1}{c}{\textbf{\begin{tabular}[c]{@{}c@{}}DMLab\\ Mazes\\ (bits / dim)\end{tabular}}} \\ \hline
\multicolumn{6}{l}{\textbf{Memory conditioned models}}                                                                                                                                                                                                                                                                                                                                                                                                                                                                                                        \\ \hline
VMA \cite{bornschein2017variational}                                & -                                                                                                       & 103.6                                                                                                      & -                                                                                                   & -                                                                          & -                                                                                                 \\
KM \cite{wu2018kanerva}                                 & -                                                                                                       & $\leq$68.3                                                                                                       & -                                                                                                   & $\leq$4.37$^{+}$                                                                       & -                                                                                                 \\
DNC \cite{graves2016hybrid}                                & -                                                                                                       & $\leq$100                                                                                                        & -                                                                                                   & -                                                                          & -                                                                                                 \\
DKM \cite{wu2018learning}                               & $\leq$75.3$^*$                                                                                                  & $\leq$77.2                                                                                                       & -                                                                                                   & $\leq$4.79$^{*}$                                                                     & \textbf{$\leq$2.75$^\dagger$}                                                                                     \\
DKM w/TSM (our impl)               & $\leq$51.84                                                                                                   & $\leq$70.88                                                                                                      & $\leq$4.15                                                                                                   & $\leq$4.31                                                                          & $\leq$2.92$^\dagger$                                                                                                 \\
Kanerva++ (ours)                    & \textbf{$\leq$41.58}
                                                                                                                                              &
                                                                                                                                                \textbf{$\leq$66.24}
                                                                                                                                                                                                                                                           &
                                                                                                                                                                                                                                                             \textbf{$\leq$3.40}                                                                                &  \textbf{$\leq$3.28}                                                       & $\leq$2.88$^\dagger$                                                                                          \\ \hline
\end{tabular}
}}
\caption{Negative test likelihood and conditional test likelihood values (lower is better). $^{*}$ was graciously provided by original authors. $^+$ estimated from \cite{wu2018kanerva} Appendix Figure 12. $^\dagger$ variadic performance due to online generation of DMLab samples.}  \label{likelihood_table}
\vskip -0.1in
\end{table}

K++ presents state-of-the-art results for memory conditioned binarized MNIST and binarized Omniglot,
and presents competitive performance for Fashion MNIST, CIFAR10 and DMLab mazes.
The performance gap on the continuous valued datasets can be explained by our
use of a simple convolutional decoder, rather than the autoregressive decoders
used in models such as PixelVAE \cite{DBLP:journals/corr/abs-1908-09948}. We
leave the exploration of more powerful decoder models to future work and note that
our model can be integrated with autoregressive decoders.

\subsection{Iterative inference}\label{genitersec}
\begin{figure}
    \begin{center}
\scalebox{1.0}{\parbox{1.0\linewidth}{
  \begin{minipage}{0.55\textwidth}
    \centerline{\includegraphics[width=\linewidth]{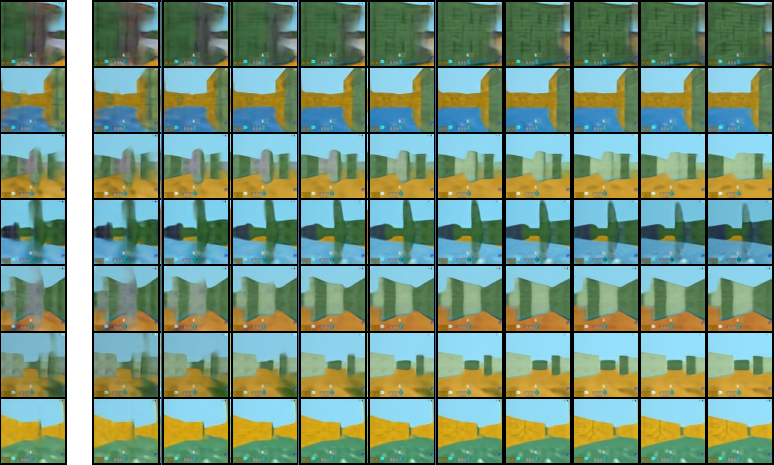}}
  \end{minipage}%
\begin{minipage}{0.425\textwidth}
    \centerline{\includegraphics[width=0.9\linewidth]{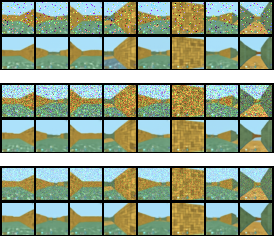}}
  \end{minipage}%
}}
\end{center}
 \caption{\emph{Left}: First column to left visualizes first random key generation. Following columns
   created by inferring previous sample through K++. \emph{Right}:
   denoising of salt \& pepper (top), speckle (middle) and Poisson
   noise (bottom).}\label{cleanupgen}
 \end{figure}
One of the benefits of K++ is that it uses the memory to
learn a more informed prior by condensing the information from an
episode of samples. One might suspect that %
based
on the dimensionality of the memory and the size of the read traces,
the memory might only learn prototypical patterns, rather than a
full amalgamation of the input episode.
This presents a problem for
generation, as described in Section \ref{sgm} , and can be  observed
in the first column of Figure \ref{cleanupgen}-\emph{Left} where the
first generation from a random key appears blurry. %
To overcome this limitation, we rely on the iterative inference %
of Kanerva memory models \citep{wu2018kanerva,wu2018learning}. By holding
the memory, $M$, fixed and repeatedly inferring the latents, %
we are able to clean-up the pattern by leveraging the
contextual information contained within the memory ($\S$ \ref{rm}).
This is visualized in the proceeding columns of Figure \ref{cleanupgen}-\emph{Left}, where we observe a slow but clear improvement in
generation quality. This property of iterative inference %
is one of the central benefits of using a memory model over a tradition solution like a VAE.
We also present results of iterative inference on more classical image noise
distributions such as salt-and-pepper, speckle and Poisson noise in Figure
\ref{cleanupgen}-\emph{Right}. For each original noisy pattern (top rows) we
provide the resultant final reconstruction after ten steps of clean-up. The
proposed K++ is robust to input noise and is able to clean-up most of
the patterns in a semantically meaningful way.

\vspace{-0.05in}
\subsubsection{Image Generations}\label{newgeneration}

\begin{figure}[H]
  \begin{center}
\scalebox{1.0}{\parbox{1.0\linewidth}{
  \begin{minipage}{0.33\textwidth}
    \centerline{\includegraphics[width=\linewidth]{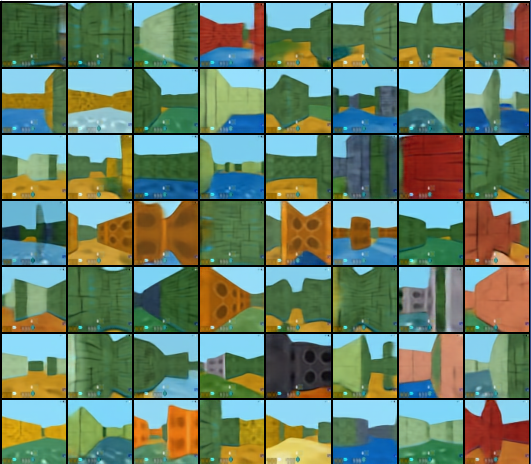}}
  \end{minipage}%
\begin{minipage}{0.325\textwidth}
    \centerline{\includegraphics[scale=0.553]{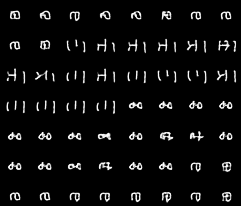}}
  \end{minipage}%
\begin{minipage}{0.33\textwidth}
  \centerline{\includegraphics[scale=0.12172]{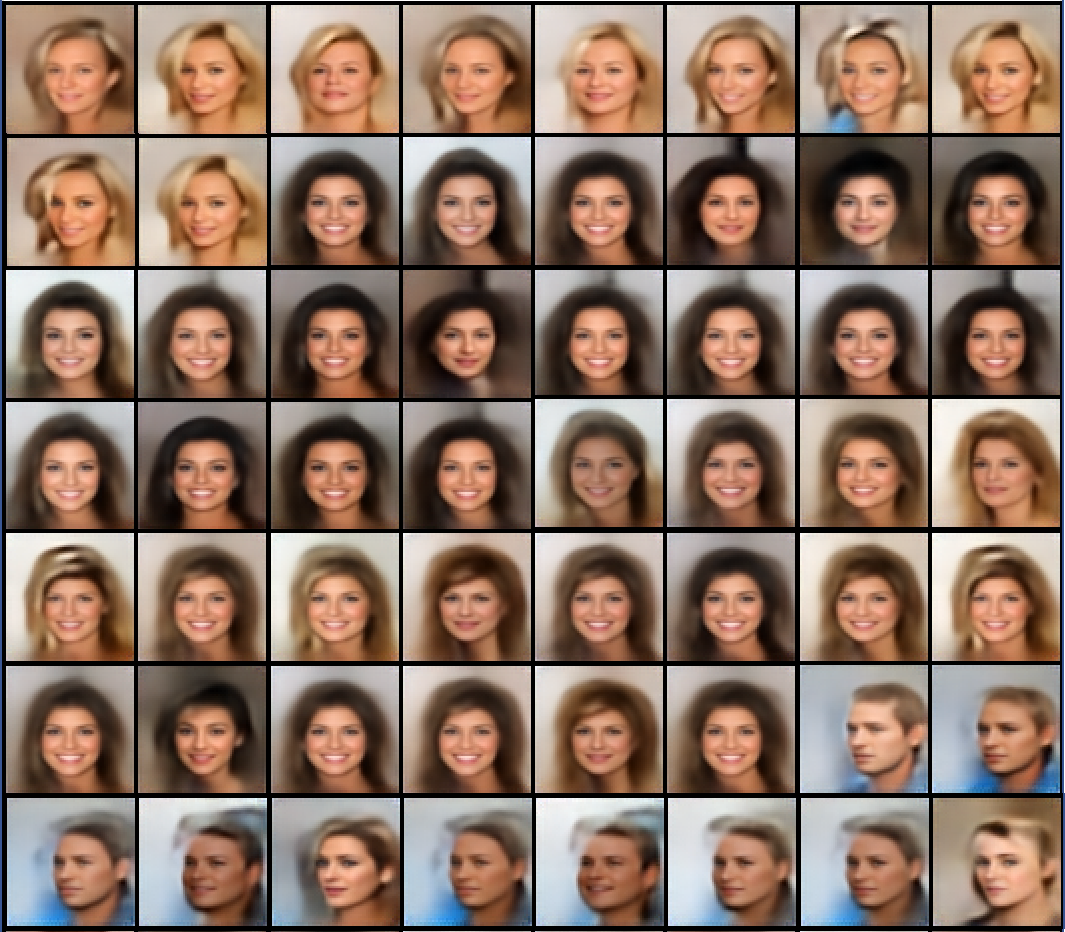}}
\end{minipage}%
}}
\end{center}
\vskip -0.1in
 \caption{\small Key perturbed generations. \emph{Left}: DMLab
   mazes. \emph{Center}: Omniglot. \emph{Right}: Celeb-A 64x64.}\label{newgenfig}
\end{figure}

\begin{figure}[H]
\begin{center}
\scalebox{1.0}{\parbox{1.0\linewidth}{
  \begin{minipage}{0.33\textwidth}
    \centerline{\includegraphics[width=\linewidth]{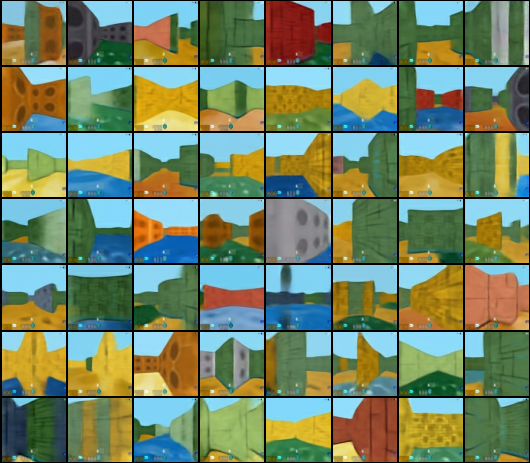}}
  \end{minipage}%
\begin{minipage}{0.325\textwidth}
    \centerline{\includegraphics[scale=0.553]{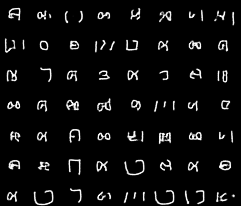}}
  \end{minipage}%
\begin{minipage}{0.33\textwidth}
  \centerline{\includegraphics[scale=0.2463]{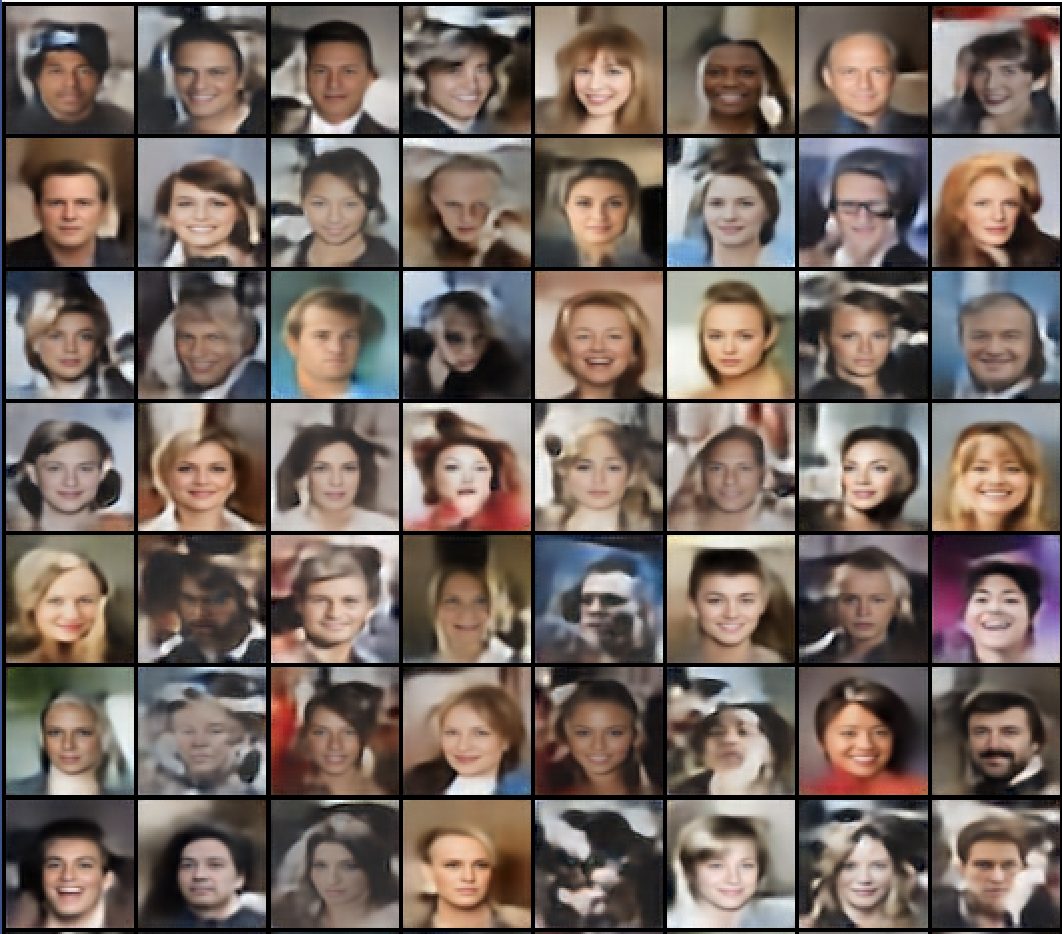}}
\end{minipage}%
}}
\end{center}
\vskip -0.1in
 \caption{\small Random key generations. \emph{Left}: DMLab
   mazes. \emph{Center}: Omniglot. \emph{Right}: Celeb-A 64x64.}\label{uncondgen}
\end{figure}

Typical VAEs
 use high dimensional isotropic Gaussian latent variables ($\geq \mathbb{R}^{16}$)
\cite{DBLP:journals/corr/BurdaGS15,kingma2014}. A well known property
of high dimensional Gaussian distributions is that most of their mass
is concentrated on the surface area of a high dimensional
ball. Perturbations to a sample in an area of valid density can easily move it
to an invalid density region \citep{arvanitidis2018latent,white2016sampling}, causing blurry or irregular
generations. In the case of K++, since the key distribution, $y_t \sim p(Y), y_t \in \mathbb{R}^3$, is
within a low dimensional space, local perturbations, $y_t + \epsilon,
\epsilon \sim N(0, 0.1)$, are likely in regions with high probability
density.
 We visualize this form of generation in Figure \ref{newgenfig} for DMLab Mazes, Omniglot and
Celeb-A 64x64, as well as the more traditional random key generations, $y_t \sim p(Y)$, in Figure \ref{uncondgen}.
\vspace{-0.1in}
\vskip -0.2in
\vskip -0.2in

Interestingly, local key perturbations of a trained DMLab Maze K++
model induces resultant generations that provide a natural traversal
of the maze as observed by scanning Figure \ref{newgenfig}-\emph{Left}, row by row, from left to right.
In contrast, the random generations of the same task (Figure \ref{uncondgen}-\emph{Left}) present a more
discontinuous set of generations. We see a similar effect for the
Omniglot and Celeb-A datasets, but observe that the locality is
instead tied to character or facial structure as shown in Figure
\ref{newgenfig}-\emph{Center} and Figure \ref{newgenfig}-\emph{Right}. Finally, in
contrast to VAE generations, K++ is able
to generate sharper images of ImageNet32x32 as shown in Appendix
\ref{vae_vs_kpp}. Future work will investigate this form of locally
perturbed generation through an MCMC lens.

\vspace{-0.05in}
\subsection{Ablation: Is Block Allocated Spatial Memory Useful?}
\vspace{-0.1in}
\begin{figure}[H]
\begin{center}
\scalebox{1.0}{\parbox{1.0\linewidth}{
\begin{minipage}{0.6\textwidth}
  \includegraphics[width=\linewidth]{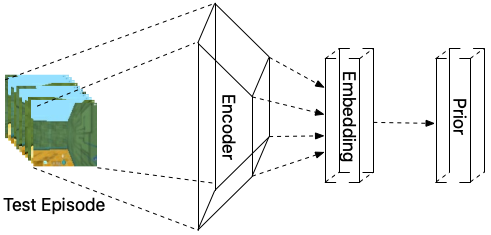}
\end{minipage}%
\hfill\vline\hfill
\begin{minipage}{0.39\textwidth}
  \includegraphics[width=\linewidth]{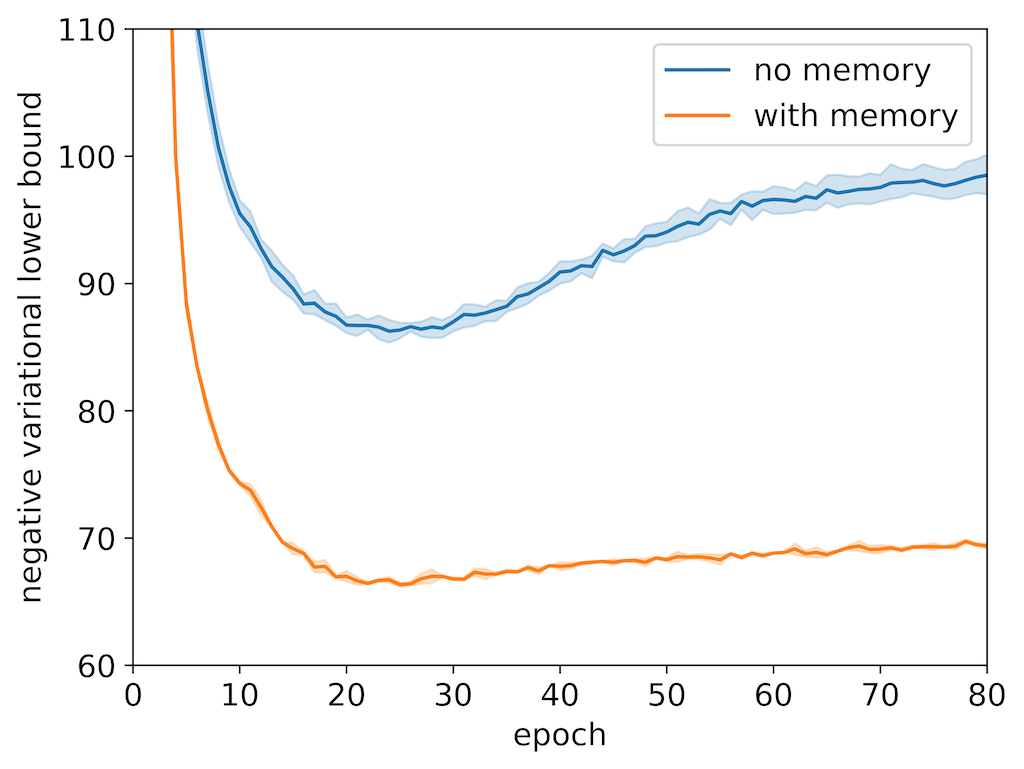}
\end{minipage}%
}}
\end{center}
\caption{\emph{Left}: Simplified write model directly produces readout prior from Equation \ref{actual_elbo} by projecting embedding $E$ via a learned network. \emph{Right}: Test
  negative variational lower bound (mean $\pm$1std).} \label{mem_useful_fig}
\end{figure}
\vspace{-0.2in}

While Figure \ref{newgenfig} demonstrates the advantage of having a
low dimensional sampling distribution and Figure \ref{cleanupgen}
demonstrates the benefit of iterative inference, it is unclear whether
the performance benefit in Table
\ref{likelihood_table} is achieved from the episodic training, model
structure, optimization procedure or memory allocation scheme. To
isolate the cause of the performance benefit, we simplify the write architecture from
Section \ref{wm} as shown in Figure
\ref{mem_useful_fig}-\emph{Left}. In this scenario, we produce the
learned memory readout, $Z$, via an equivalently sized
dense model that projects the embedding, $E$,
while keeping all other aspects the same.
We train both models five times with the exact same TSM-ResNet18 encoder,
decoder, optimizer and learning rate scheduler. As shown in Figure
\ref{mem_useful_fig}-\emph{Right}, the test conditional variational lower bound of the K++ model is \textbf{20.6 nats/image}
better than the baseline model for the evaluated binarized Omniglot dataset. This confirms that the spatial, block
allocated latent memory model proposed in this work is useful when
working with image distributions. Future work will explore this dimension
for other modalities such as audio and text.

\vspace{-0.05in}
\subsection{Ablation: episode length (T) and memory read steps (K).}
\vspace{-0.1in}

\begin{figure}[H]
\begin{center}
\scalebox{1.0}{\parbox{1.0\linewidth}{
\begin{minipage}{0.5\textwidth}
  \includegraphics[width=\linewidth]{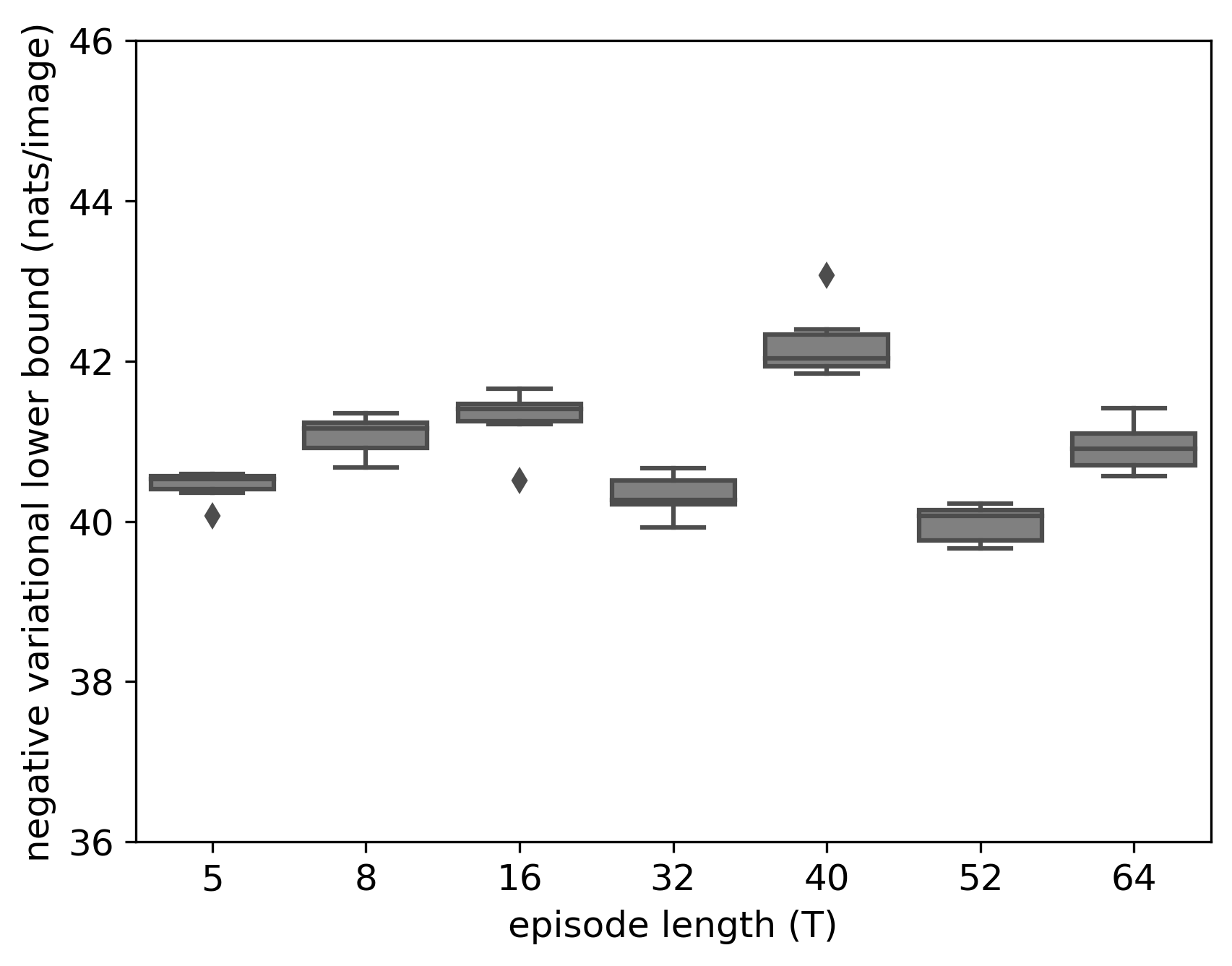}
\end{minipage}%
\begin{minipage}{0.49\textwidth}
  \includegraphics[width=\linewidth]{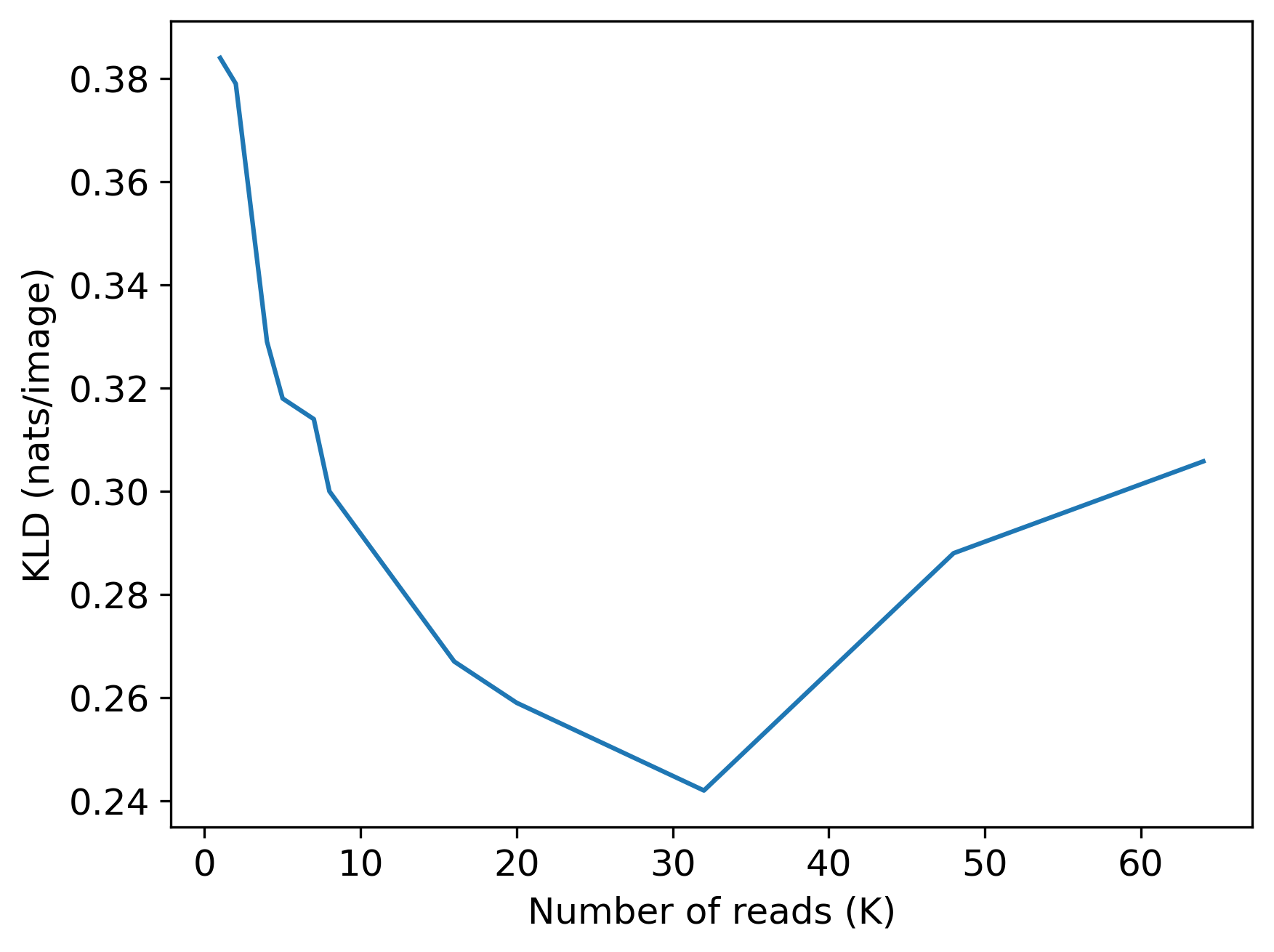}
\end{minipage}%
}}
\end{center}
\caption{Binarized MNIST. \emph{Left}: Episode length (T) ablation showing negative test
conditional variational lower bound (mean$\pm$std). We suspect the variance at $T=40$ is related to the fractional shift of the TSM-encoder. \emph{Right}: Memory read steps (K) ablation showing
test KL divergence.} \label{TKablation}
\end{figure}

To further explore K++, we evaluate the sensitivity of the model to varying
episode lengths (T) in Figure \ref{TKablation}-\emph{left} and memory read steps
(K) in Figure \ref{TKablation}-\emph{right} using the binarized MNIST dataset.
We train K++ five times (each) for episode lengths ranging from 5 to 64
and observe that the model performs within margin of error for increasing episode lengths, producing
negative test conditional variational bounds within a 1-std of $\pm 0.625$ nats/image.
This suggests that for the specific dimensionality of
memory ($\mathbb{R}^{64 \times 64}$) used in this experiment, K++ was able to
successfully capture the semantics of the binarized MNIST distribution. We
suspect that for larger datasets this relationship might not necessary hold and
that the dimensionality of the memory should scale with the size of the dataset,
but leave the prospect of such capacity analysis for future research.

While ablating the number of memory reads (K) in Figure
\ref{TKablation}-\emph{right}, we observe that the total test KL-divergence
varies by 1-std of $\pm$0.041 nats/image for a range of memory reads from 1 to 64. A
lower KL divergence implies that the model is able to better fit the approximate
posteriors $q_{\phi}(Z|X)$ and $q_{\phi}(Y|X)$ to their correspondings priors in
Equation \ref{actual_elbo}. It should however be noted that a lower
KL-divergence does not necessary imply a better generative model
\cite{DBLP:journals/corr/TheisOB15}. While qualitatively inspecting the
generated samples, we observed that K++ generated more semantically sound
generations at lower memory read steps. We suspect that the difficulty of
generating realistic samples increases with the number of disjoint reads and
found that $K=2$ produces high quality results. We use this value for all
experiments in this work.

%% file: appendix.tex
\appendix

\section{Spatial Transformer Review}\label{streview}

Indexing a matrix, $M[x:x+\Delta x, y:y+\Delta y]$, is typically a
non-differentiable operation since it involves hard cropping around an
index. Spatial transformers \citep{jaderberg2015spatial} provide a
solution to this problem by decoupling the problem into
two differentiable operands:

\begin{enumerate}
  \item Learn an affine transformation of coordinates.
  \item Use a differntiable bilinear transformation.
\end{enumerate}

The affine transformation of source coordinates, $\begin{bmatrix}i^s
  \\ j^s \end{bmatrix}$, to target coordinates, $\begin{bmatrix}i^t \\
  j^t \end{bmatrix}$ is defined as:

\begin{align}
\begin{bmatrix}i^t \\ j^t \end{bmatrix} = \begin{bmatrix} s
 \ & \ 0\ & x \\ \ 0\ &  s\ \ & y \end{bmatrix} \begin{bmatrix} i^s \\ j^s \\ 1 \end{bmatrix} = \begin{bmatrix} y_0 & 0 &\  y_1 \\
  0\ & y_0 & y_2 \end{bmatrix} \begin{bmatrix} i^s \\ j^s \\
  1 \end{bmatrix} \label{steqn}
\end{align}

Here, the affine transform, $\theta = \begin{bmatrix} s\ & \ 0\ & x \\
  \ 0\ &  s\ \ & y \end{bmatrix}$ has three learnable scalars: $\{s,
x, y\}$ which define a scaling and translation in $i$ and $j$
respectively. In the case of K++, these three scalars represent the
components of the key sample, $\{y_0, y_1, y_2\} \in \mathbb{R}^3$ as shown in
Equation \ref{steqn}. After transforming the co-ordinates (not to be confused with the
actual data), spatial transformers learn a differentiable bilinear
transform which can be interpreted as learning a differentiable mask that is
element-wise multiplied by the original data, $M$:
\vskip -0.2in
\begin{align}
  \sum_{n=-1}^J \sum_{r=-1}^J \bigg( M^c_{nr} \max (0, 1-|i^t_{nr} -
  r|) \max (0, 1-|j^t_{nr} - n|) \bigg)
\end{align}

Consider the following example where $\theta = \begin{bmatrix} 0.5\
  & \ 0\ & 0.3 \\ \ 0\ &  0.5\ \ & 0.5 \end{bmatrix}$; this
parameterization differntiably extracts the region shown in Figure
\ref{grid_imgs}-\emph{Right} from Figure \ref{grid_imgs}-\emph{Left}:
\begin{figure}[H]
  \begin{center}\scalebox{0.9}{\parbox{1.0\linewidth}{%
\begin{minipage}{0.49\textwidth}
    \centerline{\includegraphics[width=\linewidth]{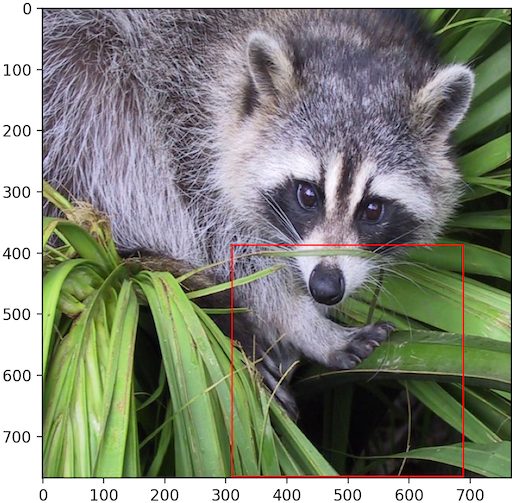}}
  \end{minipage}%
\begin{minipage}{0.49\textwidth}
    \centerline{\includegraphics[width=\linewidth]{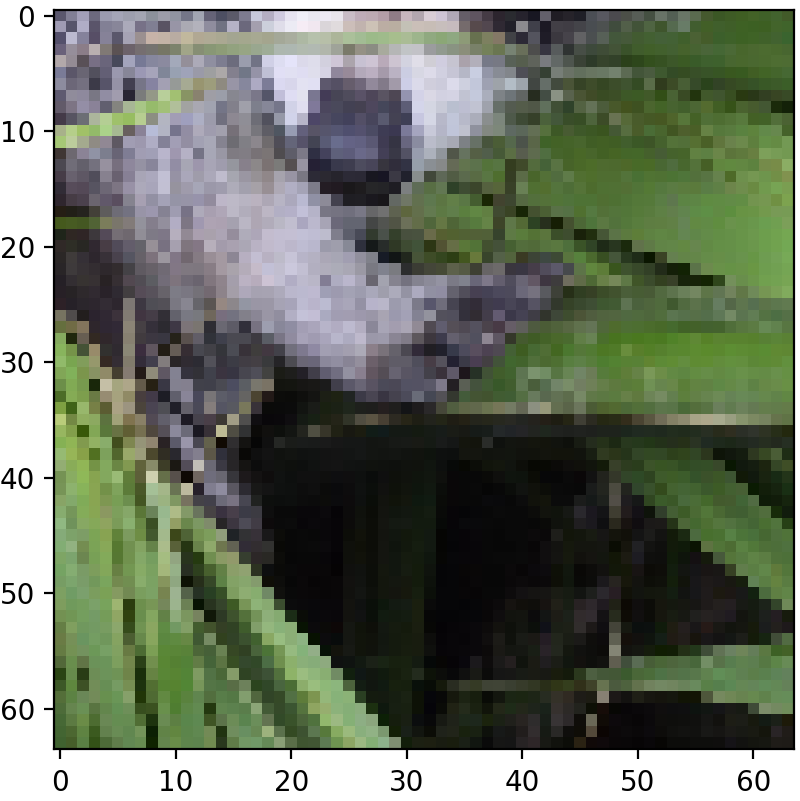}}
  \end{minipage}%
}}\end{center}
 \caption{\small Spatial transformer example. \emph{Left}: original
   image with region inlaid. \emph{Right}: extracted grid.}\label{grid_imgs}
\end{figure}

The range of values for $\{s, x, y\}$ is bound between $[-1, 1]$,
where the center of the image is $[0, 0]$. %

\section{Celeb-A Generations}\label{celeba_gen_sec}
\begin{figure}[H]
  \begin{center}\scalebox{1.0}{\parbox{1.0\linewidth}{%
\begin{minipage}{0.49\textwidth}
    \centerline{\includegraphics[width=\linewidth]{./imgs/celeba_gen2}}
  \end{minipage}%
\begin{minipage}{0.49\textwidth}
    \centerline{\includegraphics[width=\linewidth]{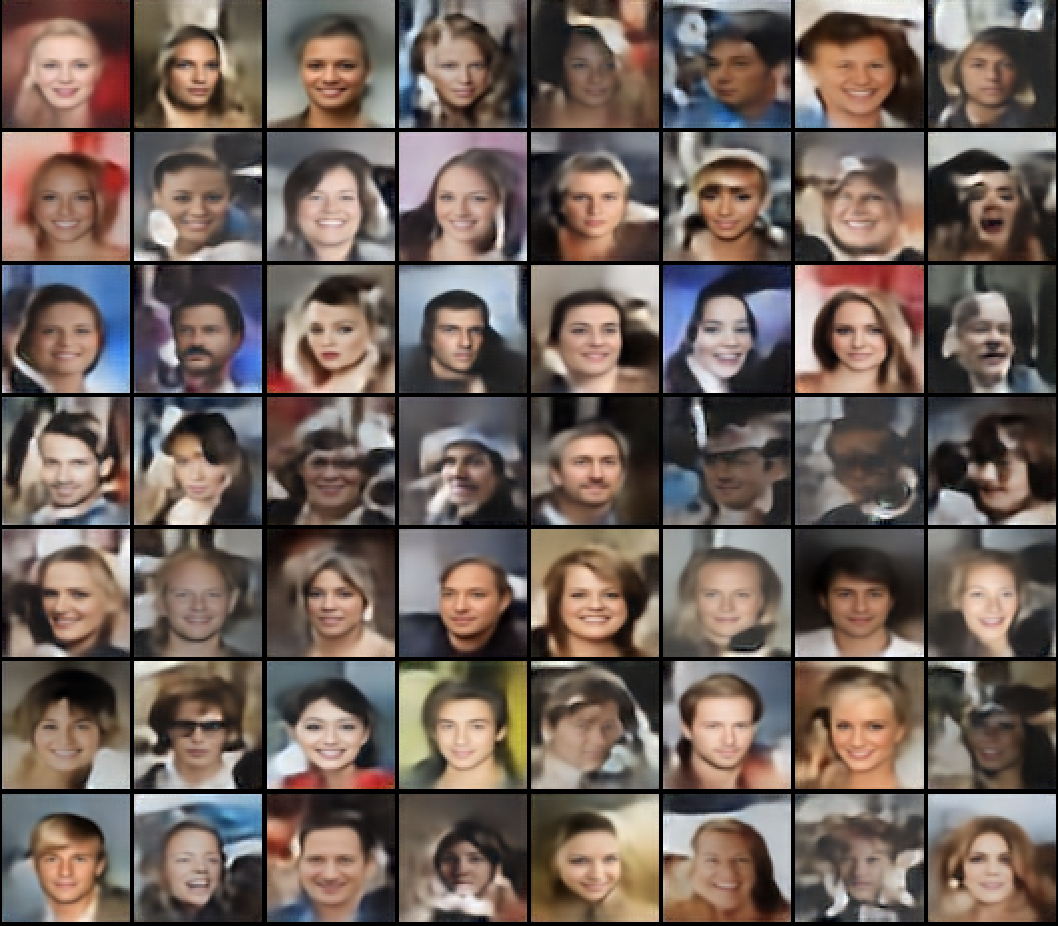}}
  \end{minipage}%
}}\end{center}
 \caption{\small Random key Celeb-A generations.}\label{celeba_gen}
\end{figure}

We present random key generations of Celeb-A 64x64, trained without
center cropping in Figure \ref{celeba_gen}.

\section{VAE vs. K++  ImageNet32x32 Generations}\label{vae_vs_kpp}

\begin{figure}[H]
  \begin{center}\scalebox{1.0}{\parbox{1.0\linewidth}{%
\begin{minipage}{0.49\textwidth}
    \centerline{\includegraphics[width=0.995\linewidth]{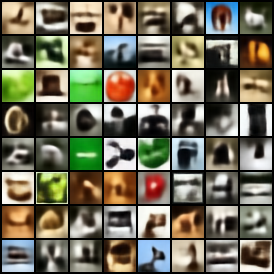}}
  \end{minipage}%
\begin{minipage}{0.49\textwidth}
    \centerline{\includegraphics[width=0.99\linewidth]{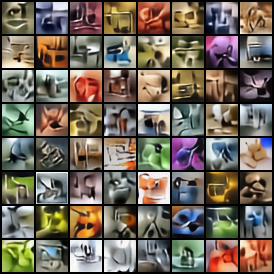}}
  \end{minipage}%
}}\end{center}
 \caption{\small ImageNet32x32 generations. \emph{Left}: VAE; \emph{Right}: K++.}\label{vae_vs_kpp_fig}
\end{figure}

Figure \ref{vae_vs_kpp_fig} shows the difference %
in generations of a standard VAE vs. K++. %
In contrast to the standard VAE
generation (Figure \ref{vae_vs_kpp_fig}-\emph{Left}), the K++
generations (Figure \ref{vae_vs_kpp_fig}-\emph{Right}) appear much
sharper, avoiding the blurry generations observed with standard VAEs.

\section{Test Image Reconstructions}\label{reconstructions}

\begin{figure}[H]
  \begin{center}\scalebox{0.70}{\parbox{1.0\linewidth}{%
  \begin{minipage}{0.49\textwidth}
    \centerline{\includegraphics[width=\linewidth]{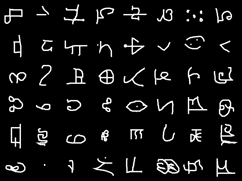}}
  \end{minipage}%
\begin{minipage}{0.49\textwidth}
    \centerline{\includegraphics[width=0.98\linewidth]{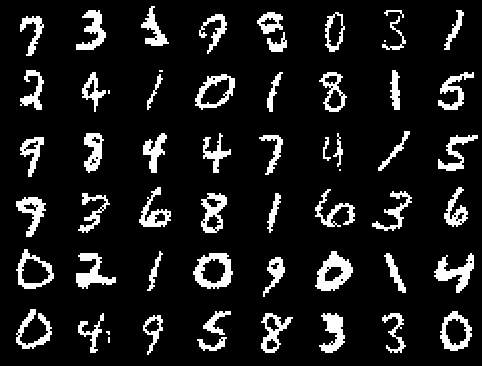}}
  \end{minipage}%

  \begin{minipage}{0.49\textwidth}
    \centerline{\includegraphics[width=\linewidth]{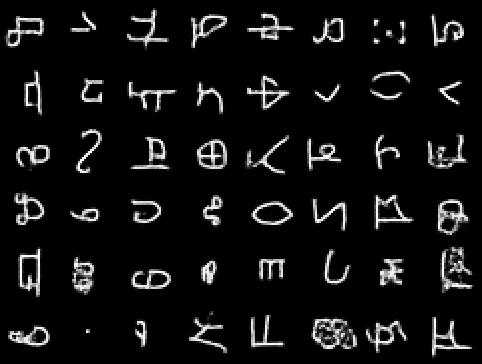}}
  \end{minipage}%
\begin{minipage}{0.49\textwidth}
    \centerline{\includegraphics[width=0.98\linewidth]{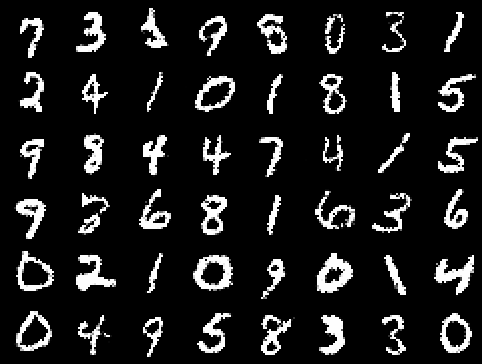}}
  \end{minipage}%
}}\end{center}
 \caption{\small Binarized test reconstructions; top row are true
   samples. \emph{Left}: Omniglot; \emph{Right}: MNIST.}\label{image_reconstr1}
\end{figure}

\begin{figure}[H]
  \begin{center}\scalebox{0.85}{\parbox{1.0\linewidth}{%
  \begin{minipage}{0.49\textwidth}
    \centerline{\includegraphics[width=\linewidth]{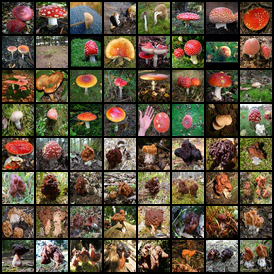}}
  \end{minipage}%
\begin{minipage}{0.49\textwidth}
    \centerline{\includegraphics[width=0.98\linewidth]{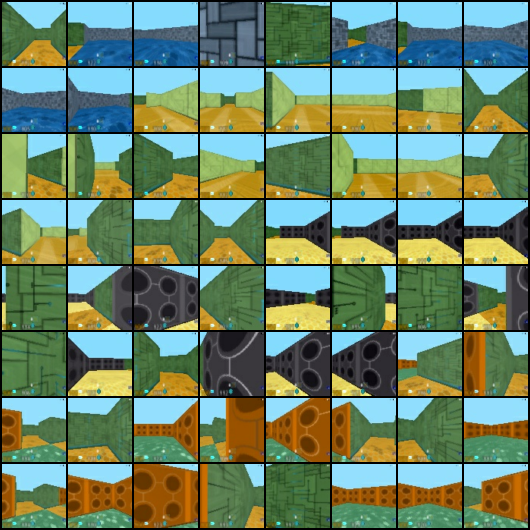}}
  \end{minipage}%

  \begin{minipage}{0.49\textwidth}
    \centerline{\includegraphics[width=\linewidth]{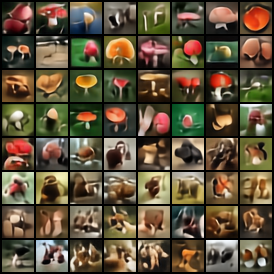}}
  \end{minipage}%
\begin{minipage}{0.49\textwidth}
    \centerline{\includegraphics[width=0.98\linewidth]{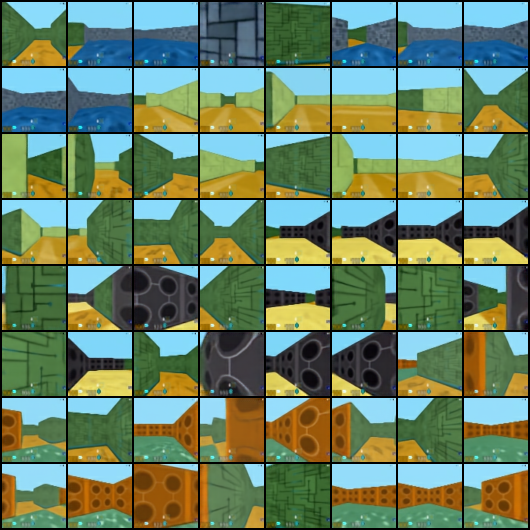}}
  \end{minipage}%
  }}\end{center}
 \caption{\small Test set reconstructions; top row are true
   samples. \emph{Left}: ImageNet64x64. \emph{Right}: DMLab Mazes.}\label{image_reconstr2}
\end{figure}

\begin{figure}[H]
  \begin{center}\scalebox{0.9}{\parbox{1.0\linewidth}{%
\begin{minipage}{\textwidth}
    \centerline{\includegraphics[width=0.99\linewidth]{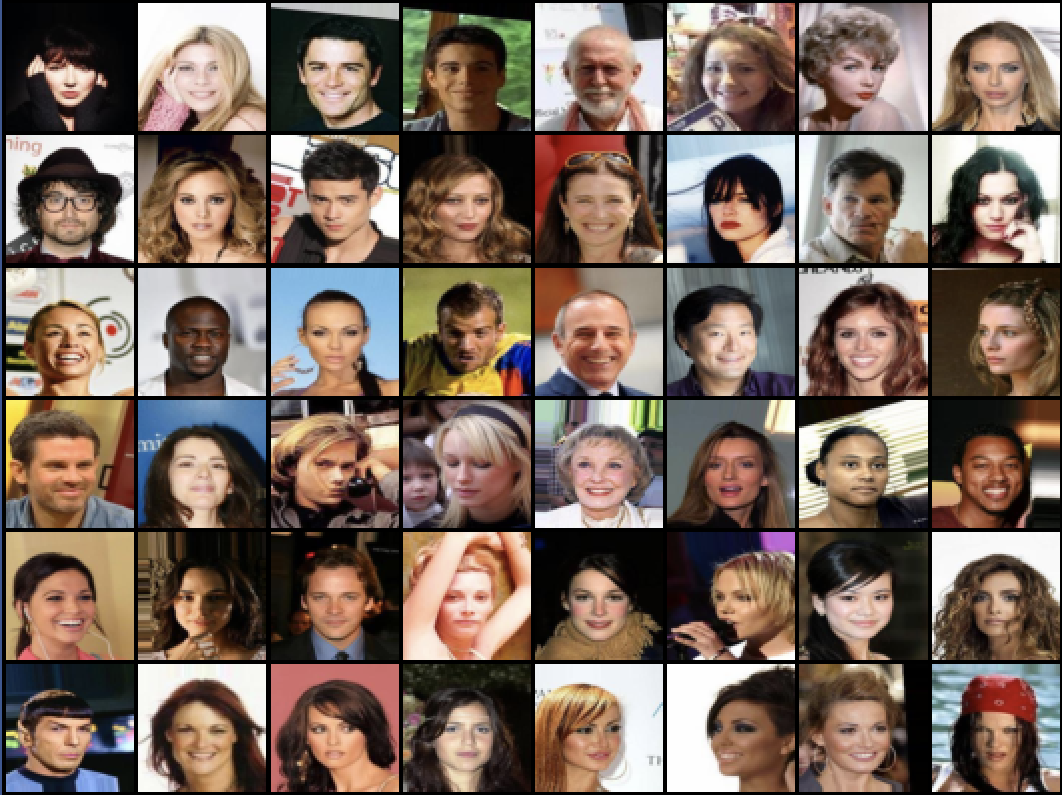}}
  \end{minipage}%

\begin{minipage}{\textwidth}
    \centerline{\includegraphics[width=0.99\linewidth]{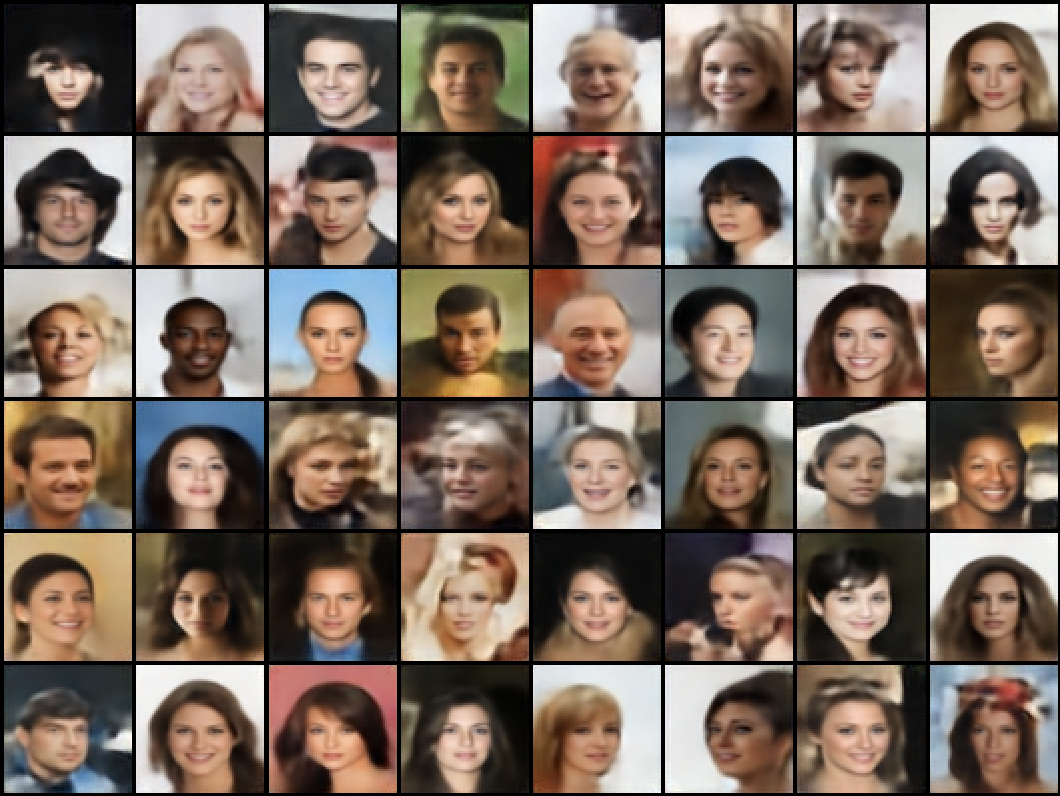}}
  \end{minipage}%
  }}\end{center}
 \caption{\small Test set reconstructions; top row are true
   samples.  Celeb-A 64x64.}\label{image_reconstr3}
\end{figure}

\section{Model Architecture \& Training} \label{appendix_arch}

\textbf{Encoder}: As mentioned in Section \ref{wm}, we use a TSM-Resnet18 encoder with
Batchnorm \citep{DBLP:conf/icml/IoffeS15} and ReLU activations \citep{DBLP:conf/icml/NairH10} for
all tasks. We apply a fractional shift of the feature maps by 0.125 as
suggested by the authors.

\textbf{Decoder}: Our decoder is a simple conv-transpose
network with EvoNormS0 \citep{DBLP:journals/corr/abs-2004-02967}
inter-spliced between each layer. Evonorms0 is similar in stride to
Groupnorm \citep{DBLP:journals/ijcv/WuH20} combined with the swish
activation function
\citep{DBLP:conf/iclr/RamachandranZL18}.

\textbf{Optimizer \& LR scheule}: We use LARS \citep{you2017large} coupled with ADAM
\citep{kingma2014adam} and a one-cycle
\citep{DBLP:journals/corr/abs-1708-07120} cosine
learning rate schedule \citep{DBLP:conf/iclr/LoshchilovH17}. A linear
warm-up of 10 epochs \citep{DBLP:journals/corr/GoyalDGNWKTJH17} is also
used for the schedule. A weight decay of $1e-3$ is used on every
parameter barring biases and the affine terms of batchnorm. Each task
is trained for 500 or 1000 epochs depending on the size of the dataset.

\textbf{Dense models}: All dense models such as our key network are simple three layer deep linear dense
models with a latent dimension of 512 coupled with spectral normalization
\citep{DBLP:conf/iclr/MiyatoKKY18}.

\textbf{Memory writer}: $f_{\theta_{mem}}$ uses a deep linear
conv-transpose decoder on the pooled embedding, $E$ with a base
feature map projection size of 256 with a division by 2 per layer. We
use a memory size of $\mathbb{R}^{3 \times 64 \times 64}$ for all the
experiments in this work.

\textbf{Learned Prior}: $p_{\theta}(Z|\hat{M}, Y)$ uses a
convolutional encoder that stacks the $K$ read traces, $\{f_{ST}(M,
y_{tk})\}_{k=1}^K$, along the channel dimension and projects it to the dimensionality of $Z$.

In practice, we observed that K++ is about 2x as fast (wall clock) compared to
our re-implementation of DKM. We mainly attribute this to not having to solve an
inner OLS optimization loop for memory inference.

\subsection{Memory creation protocol}\label{appendix_exp}

The memory creation protocol of K++ is similar in stride to that of the DKM model, given the deterministic relaxations and addressing mechanism described in Sections \ref{wm} and \ref{sgm}. Each memory, $M \sim \delta[f_{mem}(f_{enc}(X))]$, is a function of an episode of samples,
$X = \{x_t\}_{t=1}^T \in \mathcal{D}$. To efficiently optimize the conditional
lower bound in Equation \ref{actual_elbo}, we parallelize the learning objective
using a set of minibatches, as is typical with the optimization of neural
networks. As with the DKM model, K++ computes the train and test conditional
evidence lower bounds in Table \ref{likelihood_table}, by first inferring the
memory, $M \sim \delta[f_{mem}(f_{enc}(X))]$, from the input episode, followed by the read out procedure as
described in Section \ref{rm}.